\documentclass[10pt,twocolumn,a4paper,fleqn]{article}
\usepackage[a4paper,left=15mm,right=15mm,top=30mm,bottom=25mm,columnsep=10mm]{geometry}
\usepackage{graphicx}
\usepackage{times}
\usepackage{titlesec}
\usepackage{indentfirst}
\usepackage[fleqn]{amsmath}
\usepackage{fancyhdr}
\usepackage{cite}
\usepackage{enumitem}
\usepackage{etoolbox}

\renewcommand{\normalsize}{\fontsize{10pt}{11.9pt}\selectfont} 
\titleformat{\section}
  {\normalfont\large\bfseries\MakeUppercase}{\thesection.}{0.5em}{}
\titlespacing*{\section}{0pt}{2.5ex plus 0.5ex minus .2ex}{1.5ex}

\titleformat{\subsection}
  {\normalfont\normalsize\bfseries}{\thesubsection.}{0.5em}{}
\titlespacing*{\subsection}{0pt}{1.5ex}{0.5ex}

\titleformat{\subsubsection}
  {\normalfont\normalsize}{\thesubsubsection.}{0.5em}{}
\titlespacing*{\subsubsection}{0pt}{2.0ex plus .5ex minus .2ex}{0.1ex}

\setlist[description]{
  font=\itshape,
  nosep,
  leftmargin=0pt,
  labelsep=0.5em
}

\AtBeginEnvironment{table}{
  \scriptsize
}

\usepackage[small,bf]{caption}
\DeclareCaptionLabelSeparator{periodspace}{.~ }
\makeatletter
\newcommand{\affils}[1]{\def\@affils{#1}}
\renewcommand{\abstract}[1]{\def\@abstract{#1}}
\newcommand{\keywords}[1]{\def\@keywords{#1}}

\renewcommand{\@maketitle}{%
  \newpage
  \null
  \begin{center}
    {\fontsize{15pt}{15pt}\selectfont \bfseries \@title \par}
    \vskip 1.0em
    {\large \@author \par}
    \vskip 0.5em
    {\normalsize \@affils \par}
  \end{center}
  {\normalsize \noindent \textbf{Abstract:} \@abstract \par}
  \vskip 1em
  {\normalsize \noindent \textbf{Keywords:} \@keywords \par}
  \vskip 1.4em
}
\makeatother

\makeatletter
\renewenvironment{thebibliography}[1]{
  \section*{\refname}
  \normalsize
  \list{[\arabic{enumi}]}{
    \settowidth\labelwidth{[#1]}
    \leftmargin\labelwidth
    \advance\leftmargin\labelsep
    \setlength{\itemsep}{0pt}
    \setlength{\parsep}{0pt}
    \setlength{\topsep}{0pt}
    \setlength{\partopsep}{0pt}
    \usecounter{enumi}
    
  }
  
  \sloppy\clubpenalty4000\widowpenalty4000
  \sfcode`\.=1000\relax
}{
  \endlist
}
\makeatother

\fancypagestyle{firstpage}{
  \fancyhf{}
  
  \fancyfoot[C]{\footnotesize \sffamily Author's version. In Proc. of the Joint Symposium of AROB 31st and ISBC 11th (AROB-ISBC 2026), pp.~1580--1584, 2026.}
}

\title{Tracking Feral Horses in Aerial Video Using Oriented Bounding Boxes}

\author{Saeko Takizawa${}^{1}$, Tamao Maeda${}^{2}$, Shinya Yamamoto${}^{3}$, and Hiroaki Kawashima${}^{1}$}

\affils{${}^{1}$Graduate School of Information Science, University of Hyogo, Hyogo, Japan\\
(E-mail: s.takizawa.student@gmail.com, kawashima@gsis.u-hyogo.ac.jp)\\
${}^{2}$Research Center for Integrative Evolutionary Science, The Graduate University for Advanced Studies, Kanagawa, Japan\\
(E-mail: tamao@powarch.com)\\
${}^{3}$Institute for the Future of Human Society, Kyoto University, Kyoto, Japan\\
(E-mail: shinyayamamoto1981@gmail.com)\\
}
\abstract{%
The social structures of group-living animals such as feral horses are diverse and remain insufficiently understood, even within a single species. To investigate group dynamics, aerial videos are often utilized to track individuals and analyze their movement trajectories, which are essential for evaluating inter-individual interactions and comparing social behaviors. Accurate individual tracking is therefore crucial. In multi-animal tracking, axis-aligned bounding boxes (bboxes) are widely used; however, for aerial top-view footage of entire groups, their performance degrades due to complex backgrounds, small target sizes, high animal density, and varying body orientations. To address this issue, we employ oriented bounding boxes (OBBs), which include rotation angles and reduce unnecessary background. Nevertheless, current OBB detectors such as YOLO-OBB restrict angles within a 180$^{\circ}$ range, making it impossible to distinguish head from tail and often causing sudden 180$^{\circ}$ flips across frames, which severely disrupts continuous tracking. To overcome this limitation, we propose a head-orientation estimation method that crops OBB-centered patches, applies three detectors (head, tail, and head-tail), and determines the final label through IoU-based majority voting. Experiments using 299 test images show that our method achieves 99.3\% accuracy, outperforming individual models, demonstrating its effectiveness for robust OBB-based tracking.
}


\begin{document}

\maketitle
\thispagestyle{firstpage}

\section{Introduction}

The social structures of group-living animals, such as feral horses, are diverse and highly complex \cite{maeda2021aerial}. Variations are observed even within the same species; for instance, some groups are maintained for over a year, while others change within about a month. However, many factors underlying these social variations remain unclear. In recent years, to elucidate these mechanisms, quantitative analysis based on trajectory data obtained from individual tracking in drone-captured aerial video of entire groups has been widely performed \cite{ozogany2023fine}. Such trajectory data is utilized to analyze interactions based on inter-individual distances and to quantify travel distances at both the individual and group levels.
Therefore, high-precision individual tracking is indispensable for ensuring the reliability of these analyses.

In general object tracking, detection results are associated based on movement distance or cosine similarity between frames. In this process, axis-aligned bounding boxes (bbox) are often used; however, they are not necessarily suitable for the video data addressed in this study. In particular, our aerial video is characterized by very small individual sizes relative to the entire image due to high shooting altitude, the inclusion of many shadows and depressions, dense clustering of individuals, and each individual facing a different direction. In such video, false positives in areas where no horses are present, such as shadows and depressions, and false negatives where horses are present but not detected, occur.
Therefore, in this study, to leverage the characteristics of video captured from above, we use oriented bounding boxes (OBBs) \cite{wang2019mask}, which consider the rotation angle of the object when viewed from above. By using OBB, it becomes possible to enclose each horse while minimizing background inclusion, even when the animals are rotating; thus, it is objectively expected that false positives can be reduced. However, widely used OBB detectors adopt the range of 0$^{\circ}$ to 180$^{\circ}$ for implementation simplicity and data availability. To address this, we propose a multi-stage detection model that estimate the head direction by detecting the head and tail positions after individual detection as a method for estimating the head direction. This approach maintains temporal continuity of each individual's OBB angle and aims to improve tracking stability.

\section{Related work}
To address the tracking of group-living feral horses, we first describe the Tracking-by-Detection framework, a general approach in multiple object tracking (MOT) that performs object detection prior to tracking.

\subsection{YOLO-based object detection}
YOLO \cite{khanam2024yolov11} is a representative object detection algorithm. Specifically, YOLO11 \cite{khanam2024yolov11} is a detection model that predicts object positions and classes across three scales---large, medium, and small---based on feature maps extracted by a convolutional neural network (CNN). During inference, the input is an image, and the outputs are the bbox center coordinates, width, height, confidence score, and class probabilities. For each pixel in the feature maps integrated from each scale, the model directly regresses the objectness, bbox coordinates ($x, y, w, h$), and class probabilities. The distances from the center of each pixel to the four sides (top, bottom, left, right) of the bbox are learned as a probability distribution using distribution focal loss (DFL). These distances are regressed as continuous values based on this probability distribution, from which the center coordinates, width, and height of the bbox are calculated. Class probabilities are computed by applying a sigmoid function and multiplying the result by the objectness score. Finally, predictions with confidence scores below a threshold are excluded, and duplicate bboxes are removed using Non-Maximum Suppression (NMS) to produce the final output.

\subsection{YOLO OBB model}
In OBB-based object detection, the model takes an image as input and outputs the object position (center coordinates, width, height) along with its orientation (angle or four-point coordinates), class, and confidence score. 
This is particularly effective for top-down imagery such as aerial photography and satellite images.
Within the Ultralytics framework, YOLO OBB models \cite{YOLO_OBB} have been available since YOLOv8 \cite{yaseen2408yolov8}.
However, in the YOLO OBB implementations, the rotation range is constrained to a 180-degree interval, a limitation that is addressed in this paper.

\subsection{Multiple object tracking methods}
Simple Online and Realtime Tracking (SORT) \cite{bewley2016simple} is a representative method for multiple object tracking. SORT first employs a Kalman filter to predict the bbox state of each existing track in the next frame. A cost matrix is then constructed by computing the Intersection over Union (IoU) between the predicted bboxes and the detected bboxes. The Hungarian algorithm is used to associate predictions with detections by finding the assignment that minimizes the total cost. However, SORT has a limitation in that object IDs are lost and reinitialized when occlusion occurs. Many SORT-based tracking methods overcome this issue by extending a cost matrix that incorporates additional information (e.g., appearance features~\cite{wojke2017simple}). Using enhanced cost, predictions and detections are associated more reliably, enabling the tracker to maintain consistent IDs even when an object reappears after being occluded. This paper mainly focuses on the detection stage which can be integrated with these tracking methods.

\subsection{Pose estimation methods}
Representative pose estimation methods include SLEAP \cite{pereira2022sleap}, DeepLabCut \cite{mathis2018deeplabcut}, and the YOLO Pose model \cite{YOLO_POSE}. In these approaches, images or videos are fed into a model trained to recognize the keypoints of the target animals and their spatial relationships. The model outputs the coordinates and confidence scores of the keypoints for each individual. Furthermore, SLEAP and DeepLabCut support not only pose estimation but also individual tracking based on the estimated poses. Pose estimation applied to our videos produced erroneous results, such as redundant detections and multiple overlapping pose estimates for a single individual. We therefore adopted OBB-based detection instead of pose estimation.

\section{Proposed Method}
\begin{figure}[t]
\begin{center}
\includegraphics[width=\columnwidth]{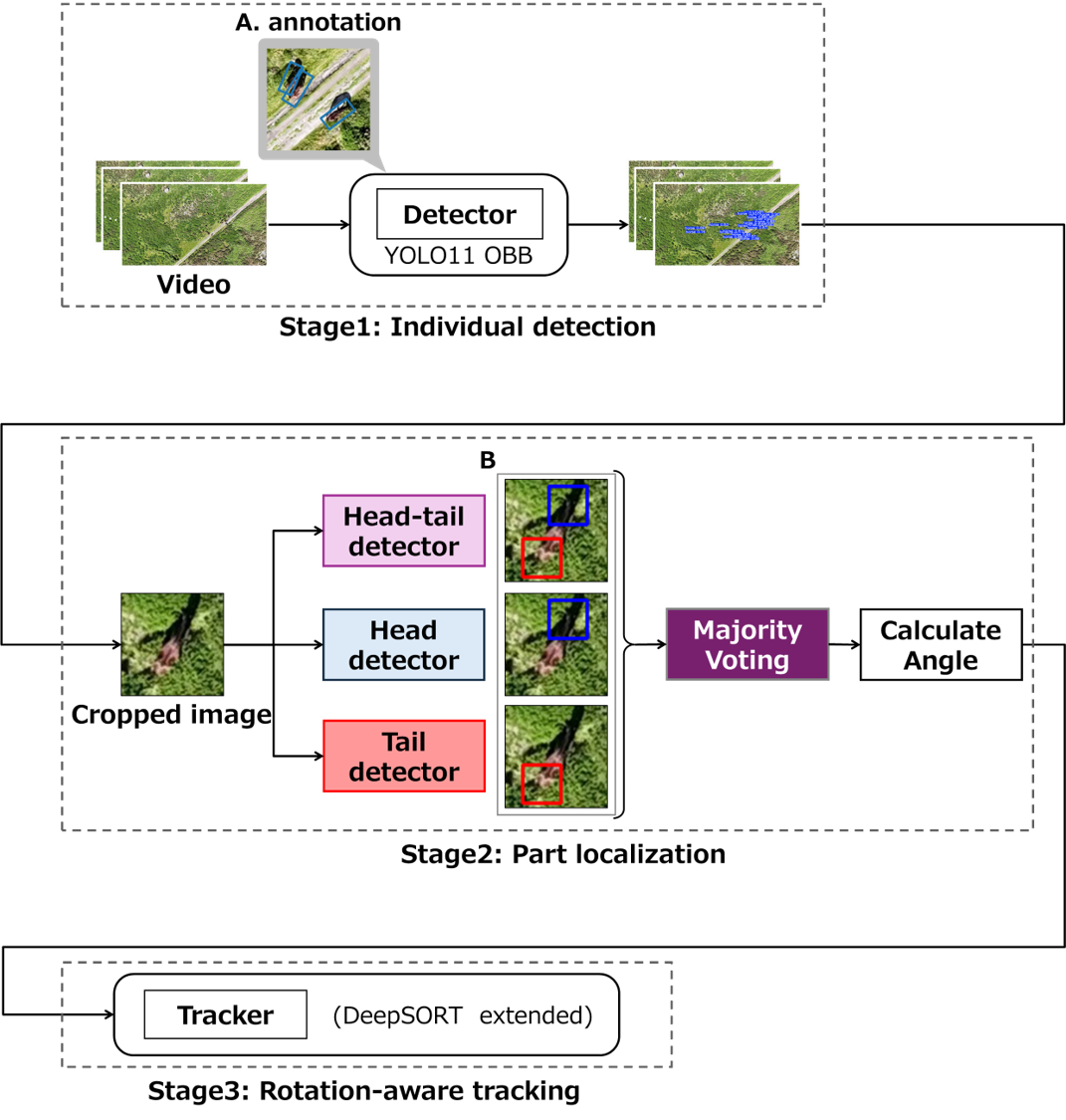}
\caption{\label{approach} Overview of the proposed method which consists of (1) individual detection, (2) part localization, and (3) rotation-aware tracking}
\end{center}
\end{figure}

Figure~\ref{approach} illustrates an overview of the proposed method which consists of (1) individual detection, (2) part localization, and (3) rotation-aware tracking.
Our method processes each frame of the video sequence as follows. Let $F_t$ denote the frame at time $t$. First, we apply OBB-based object detection to the entire frame $F_t$ and estimate the head and tail positions for each detected individual. To improve the robustness of part localization, we employ three types of detectors and determine the final head and tail coordinates through majority voting among their outputs. Based on the estimated body part positions, we compute the rotation angle in the range of 0$^{\circ}$ to 360$^{\circ}$. Finally, this angular information is provided to the tracker to perform individual tracking.

\subsection{Detection}

\subsubsection{Full-frame OBB detection of individuals}
We perform OBB detection on the entire frame $F_t$. As the detector, we employ a YOLO11m-OBB model fine-tuned using a dataset comprising 80 training images, 10 validation images, and 11 test images, based on OBB annotations as illustrated in Fig.~\ref{approach} (cropped image A).

\subsubsection{Body part detection for each individual}\label{sec:Bodypart}
Based on the OBB information of each individual detected in the previous step (Stage 1 in Fig. 1), we extract the box coordinates ($x_1, y_1, x_2, y_2$) from the entire frame $F_t$. To maintain a fixed aspect ratio for the input images to each part detector, we define a square region whose side length is the larger of the width $w$ and height $h$, anchored at the top-left coordinates ($x_1, y_1$), and crop the image. For the cropped square images, we employ three models for body part detection: the \textit{Head-Tail Detector}, the \textit{Head Detector}, and the \textit{Tail Detector}. Each model is trained for different classes: the Head-Tail Detector for both head and tail, the Head Detector for the head only, and the Tail Detector for the tail only. Combining models with different class configurations improves robustness, as relying on a single model may result in missed or incorrect detections. Each model was implemented by fine-tuning a pre-trained YOLO11m-detection model. For training, we used a dataset annotated with head and tail positions for each individual, as shown in Fig.~\ref{approach} (cropped images B). The dataset included 2,207 training images, 275 validation images, and 299 test images. The head coordinate was defined as the midpoint between the left and right ears, and the tail coordinate was set at the root of the tail. To ensure a consistent scale of body parts regardless of individual size or camera distance, each part’s bbox was defined as a square centered on corresponding part coordinate, occupying 15\% of the cropped image area.

\subsubsection{Head-tail estimation using majority voting}\label{sec:majority-voting}
\begin{figure}[t]
\begin{center}
\includegraphics[width=\columnwidth]{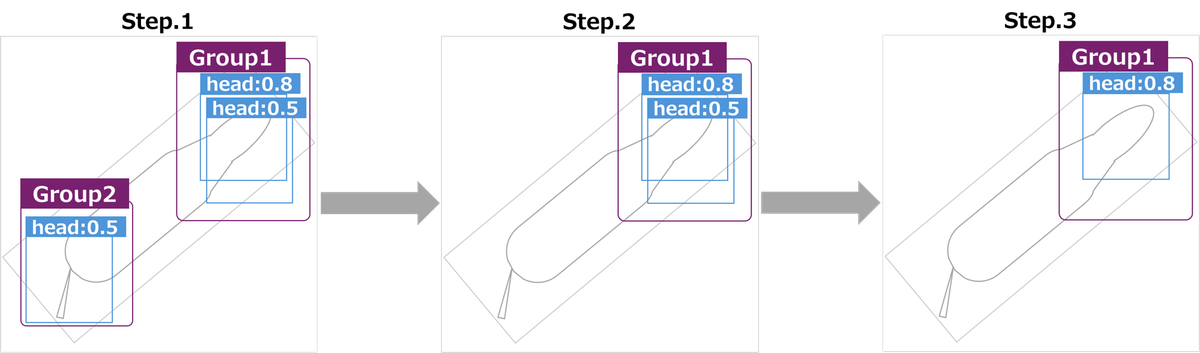}
\caption{\label{fig:voting_process} Overview of the majority voting process}
\end{center}
\end{figure}

An overview of the majority voting process is illustrated in Fig.~\ref{fig:voting_process}, which depicts an example processing flow for the head class (indicated by blue bboxes).
First, the detected part regions obtained in Sec.~\ref{sec:Bodypart} are separated by class (head and tail). Next, within each class, bboxes with an IoU of 0.3 or higher are regarded as detections referring to the same location and are clustered into a single bbox group (Fig.~\ref{fig:voting_process}, Step 1). Within each group, the bbox with the highest confidence score is defined as the \textit{group center}. After the grouping process is completed for all detections, the number of bboxes (votes) within each group is compared, and the group with the highest number of votes is selected as the final candidate (Fig.~\ref{fig:voting_process}, Step 2). In the case where multiple groups share the highest vote count, the confidence scores of their respective \textit{group centers} are compared, and the group with the higher score is adopted as the final group. Finally, the center coordinates of the \textit{group center} in the selected group are defined as the final position of the body part (Fig.~\ref{fig:voting_process}, Step 3).

It should be noted that if voting were performed solely based on the distance between the midpoint of the OBB's shorter edge and the center of each part bbox, accidental nearby false detections (outliers) could also be counted as votes. In contrast, by performing IoU-based grouping as in the proposed method, spatial consistency can be prioritized---ensuring that multiple detectors agree on the same location---which enables the effective exclusion of isolated false positives.

\subsubsection{Rotation angle calculation}\label{sec:AngleCal}
When the head is detected, we first compute the vectors from the center of the OBB to the midpoints of its short edges. We then calculate the dot product between each of these vectors and the vector from the OBB center to the center of the detected head bbox. The short-edge direction that produces a positive dot product is selected as the head orientation vector. Conversely, if the head is not detected and only the tail is available, we similarly compute the dot products using the vector toward the estimated tail position. In this case, the short-edge direction yielding a negative dot product (i.e., the direction opposite to the tail) is assigned as the head orientation vector.
Finally, the rotation angle of the OBB is derived from the angle of the head direction.

\subsection{Tracking}
While the primary focus of this study is on detection, this section illustrates how the 360-degree orientation representation obtained in Sec.~\ref{sec:AngleCal} can be integrated into an individual tracking framework. Therefore, we adopt a relatively simple SORT-based method; specifically, we employ an extended version of the DeepSORT Kalman filter adapted for OBBs. In the original DeepSORT, the state vector is defined by the bbox center coordinates $(x, y)$, aspect ratio $\gamma$, height $h$, and their relative velocities. However, our approach requires accounting for the rotation of the OBBs. Therefore, we remove the aspect ratio $\gamma$, height $h$ and their relative velocities from the state space and additionally include $\sin\theta$ and $\cos\theta$ as new parameters to avoid angular discontinuity. Angular velocity is omitted because horses rarely undergo abrupt rotations. Consequently, the state vector $\mathbf{x}$ for each tracking target is defined as a six-dimensional vector, as shown in Eq.~(\ref{eq:deepsort_extended}).
\begin{equation}
    \mathbf{x} = [\mathit{x}, \mathit{y}, \sin \theta, \cos \theta, \dot{\mathit{x}}, \dot{\mathit{y}}]^\top
\label{eq:deepsort_extended}
\end{equation}

The Kalman filter-based tracking process consists of two stages: prediction and update. First, the state vector $\mathbf{x}_{\mathit{t}}$ at time $t$ is predicted from the state vector $\mathbf{x}_{\mathit{t}-\mathrm{1}}$ using the state transition model given by Eq.~(\ref{eq:dynamics}).
\begin{equation}
    \mathbf{x}_{\mathit{t}} = \mathit{F}\mathbf{x}_{\mathit{t-\mathrm{1}}} + \mathbf{w}_{\mathit{t}} , \qquad \mathbf{w}_{\mathit{t}} \sim \mathit{N}(\mathbf{0}, \mathit{Q_t})
    \label{eq:dynamics}
\end{equation}
Here, $Q_t$ denotes the process noise covariance matrix, and $F$ represents the state transition matrix.
Following SORT and the original DeepSORT, the proposed method assumes a constant-velocity model for the positional coordinates in the state transition model, while the angular components ($\sin \theta, \cos \theta$) are modeled to remain unchanged from the previous time step. Consequently, the state transition matrix $F$ is defined as in Eq.~(\ref{eq:deepsort_extended_F}).
\begin{equation}
 F = \begin{bmatrix}
  I_2 & O_2 & \Delta t I_2 \\
  O_2 & I_2 & O_2 \\
  O_2 & O_2 & I_2
 \end{bmatrix}
 \label{eq:deepsort_extended_F}
\end{equation}

As in the original DeepSORT, the observation model used to project the predicted state vector into the observation space defines the observation vector $\mathbf{z}_{\mathit{t}}$ as in Eq.~(\ref{eq:observationmodel}).
\begin{equation}
\begin{split}
\mathbf{z}_{\mathit{t}} &= \mathit{H}\mathbf{x}_{\mathit{t}} + \mathbf{v}_{\mathit{t}} , \qquad \mathbf{v}_{\mathit{t}} \sim \mathit{N}(\mathbf{0}, \mathit{R}) \\
\label{eq:observationmodel}
    H &= \begin{bmatrix} I_4 &  0
    \end{bmatrix} 
\end{split}
\end{equation}
Here, $H$ is the observation matrix, and $\mathbf{v}_{\mathit{t}}$ represents the observation noise following a zero-mean distribution with the observation noise covariance matrix $R$.
In the proposed method, since the state vector has been extended to include angular components, the observation vector is defined as shown in Eq.~(\ref{eq:measurement_vec}).
\begin{equation} 
\mathbf{z}_{\mathit{t}} = [\mathit{x}, \mathit{y}, \sin \theta, \cos \theta]^\top 
\label{eq:measurement_vec} 
\end{equation}

Finally, in the update step, the final estimate $\hat{\mathbf{x}}_{t}$ at time $t$ is obtained by adding the product of the Kalman gain $K_t$ and the residual between the actual measurement $\hat{\mathbf{z}}_{\mathit{t}}$ and the predicted measurement $\mathbf{z}_{\mathit{t}}$ to the predicted state $\mathbf{x}_{\mathit{t}}$.
\begin{equation}
    \hat{\mathbf{x}}_{t} = \mathbf{x}_{\mathit{t}} + \mathit{K_t}(\hat{\mathbf{z}}_{t} - \mathbf{z}_{\mathit{t}})
\end{equation}
Additionally, since $\sin \theta$ and $\cos \theta$ are estimated independently in this update step, a normalization process is applied to ensure their sum of squares equals 1, thereby maintaining geometric consistency.

\section{Experiments}
\subsection{Evaluation of Head Detection Accuracy}
In this evaluation, we compared the head detection accuracy of the Head-Tail Detector, the Head Detector, the Tail Detector, and the proposed method. We used 299 test images that were excluded from the training phase. These include three primary terrain types: green vegetation, rocky areas, and brown soil. For the evaluation of individual models, the final head determination was performed by applying the grouping process as described in Sec.~\ref{sec:majority-voting}. A detection was considered correct if the IoU between the estimated head bbox and the ground truth head bbox was 0.3 or higher.

Table~\ref{tab:exp1} shows the comparison results between the proposed method and the Head-Tail, Head, and Tail Detectors.
\begin{table}[t]
\caption{Comparison results of Experiment 1.}
\label{tab:exp1}
\begin{center}
\begin{tabular}{|l|l|}\hline
    Model   &   Head Det. Accuracy  \\\hline
    Proposed Method &  99.3 \% (297/299) \\\hline
    Head-tail Detection model & 99.0 \% (296/299)\\\hline
    Head Detection model & 98.0 \% (293/299)\\\hline
    Tail Detection model & 98.0 \% (293/299) \\\hline
\end{tabular}
\end{center}
\end{table}
The proposed majority-voting method achieved an accuracy of 99.3\%, the highest among the four approaches. The Head-Tail Detector achieved the second-highest accuracy at 99.0\%, while the Head Detector and the Tail Detector, each trained on a single class, achieved 98.0\%. These results demonstrate the effectiveness of combining three models with different class configurations and integrating their outputs through majority voting, rather than relying on any single model alone.

\begin{figure}[t]
\begin{center}
\includegraphics[width=\columnwidth]{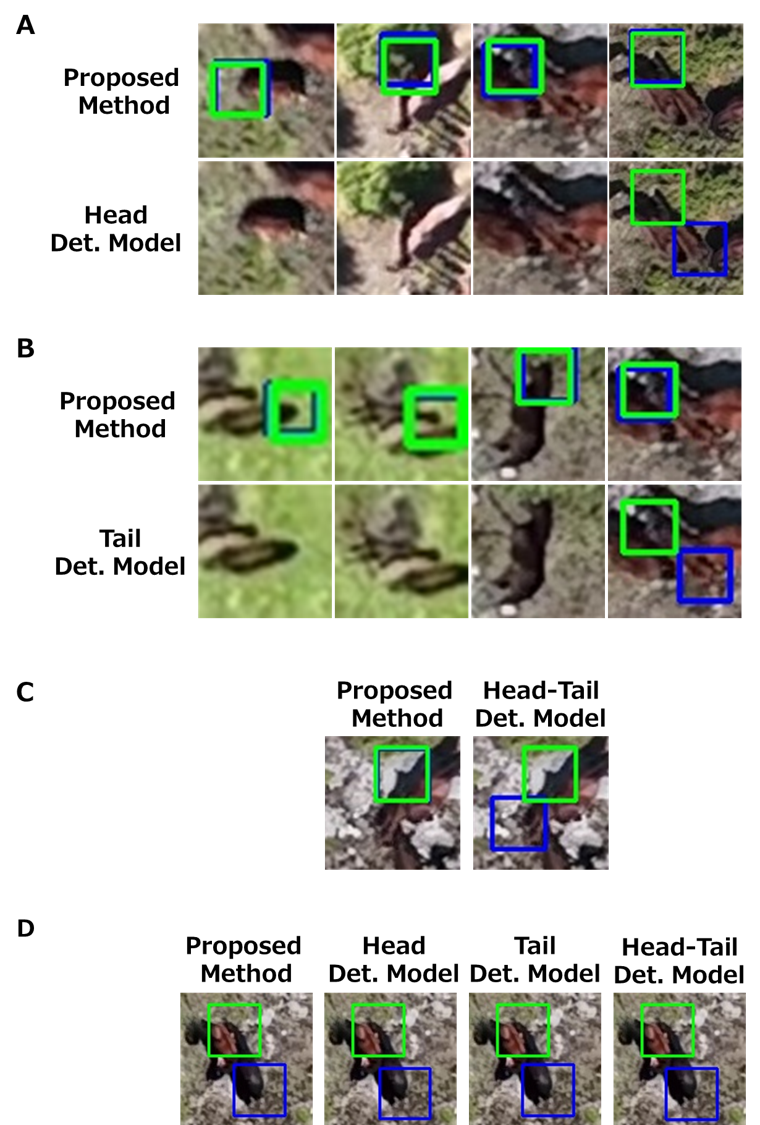}
\caption{\label{fig:result}Qualitative comparison of head estimation.}
\end{center}
\end{figure}

Furthermore, to qualitatively assess the effectiveness of the proposed method, visualization results are presented in Fig.~\ref{fig:result}.
In the figures, ground-truth head bboxes are shown in light green, while the estimated results are indicated in blue. Figs.~\ref{fig:result}A, \ref{fig:result}B, and \ref{fig:result}C present cases in which individual models fail to estimate the head position, whereas the proposed method succeeds. In Figs.~\ref{fig:result}A and \ref{fig:result}B, even when the Head Detector or Tail Detector, each trained on a single class, fails to provide a detection, the majority voting process successfully recovers the correct head location by integrating the outputs of the remaining models. In Fig.~\ref{fig:result}C, although the Head-Tail Detector produces a false positive when used alone, the proposed method correctly determines the head position by leveraging predictions from the other two detectors. Conversely, Fig.~\ref{fig:result}D illustrates a failure case for the proposed method. In images where a mare and foal are in close proximity, accurate body part estimation can be challenging due to individual occlusions and shape similarities between the horses.

\subsection{Qualitative evaluation of tracking}
\begin{figure}[t]
\begin{center}
\includegraphics[width=\columnwidth]{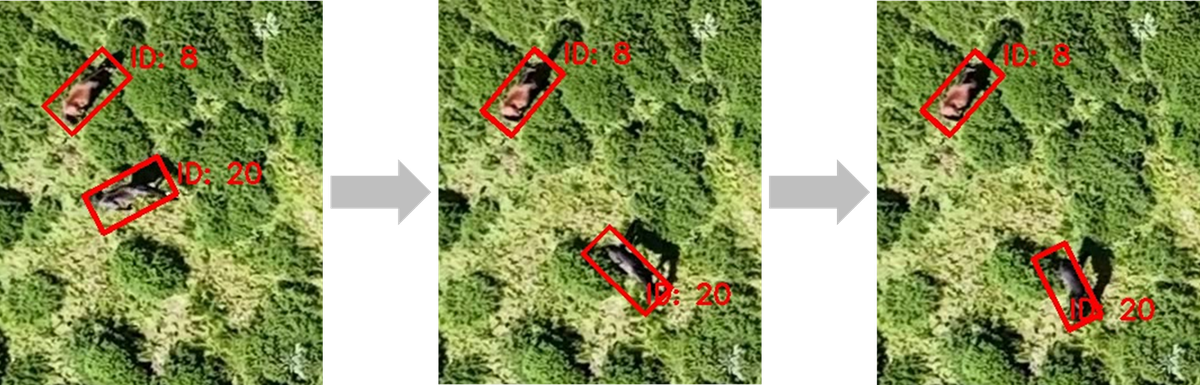}
\caption{\label{fig:result_tracking} Qualitative tracking results}
\end{center}
\end{figure}

Figure~\ref{fig:result_tracking} shows the individual tracking results using the extended DeepSORT with the orientation angles estimated by the proposed method. 
    For ID 20, it can be observed that the OBB orientation remains stable, even when the moving direction changes. These results demonstrate that the proposed method is effectively applicable to the extended DeepSORT framework.

\section{Conclusion}
In this study, we proposed a heading-aware multi-stage detection framework based on OBBs for tracking individual feral horses in aerial videos. The proposed method employs three types of detection models---the Head Detector, Tail Detector, and Head-Tail Detector---and estimates the head orientation of each individual by integrating their outputs through a majority voting scheme. The estimated orientation information is then utilized in the tracking phase.
Experimental results demonstrated that the proposed majority voting integration improves head estimation accuracy compared to using individual detection models alone. However, challenges remain when applying the method to the tracking phase. In particular, we observed cases in which errors in per-frame head estimation were directly propagated to the Kalman filter state updates, resulting in identity (ID) switches. As future work, we plan to explore the adoption of alternative tracking frameworks and to further evaluate the effectiveness of OBB-based representations in tracking.

\section*{acknowledgement}
This work was supported by JSPS KAKENHI Grant Number JP21H05302 and JP24H01432.

\end{document}